\documentclass[10pt,twocolumn,letterpaper]{article}

\usepackage[pagenumbers]{iccv} 

\usepackage{booktabs}       
\usepackage{amsfonts}       
\usepackage{nicefrac}       
\usepackage{microtype}      
\usepackage{kotex}
\usepackage{adjustbox}
\usepackage{multirow}
\usepackage{rotating}
\usepackage{bbm}
\usepackage{amsmath}
\usepackage{color, colortbl}
\usepackage{soul}
\usepackage{subcaption}
\usepackage{pifont}
\usepackage{array}
\usepackage{enumitem}

\usepackage{wrapfig}

\newcommand{\cmark}{\ding{51}}%
\newcommand{\xmark}{\ding{55}}%

\newcommand{\figref}[1]{Fig.~\ref{#1}}
\newcommand{\tabref}[1]{Tab.~\ref{#1}}
\newcommand{\secref}[1]{Sec.~\ref{#1}}

\newcommand{\equref}[1]{Eq.~(\ref{#1})}

\definecolor{iccvblue}{rgb}{0.21,0.49,0.74}
\usepackage[pagebackref,breaklinks,colorlinks,allcolors=iccvblue]{hyperref}

\usepackage{hyperref}
\usepackage{url}
\usepackage{booktabs}       
\usepackage{amsfonts}       
\usepackage{nicefrac}       
\usepackage{microtype}      
\usepackage{xcolor}         
\usepackage{kotex}
\usepackage{adjustbox}
\usepackage{multirow}
\usepackage{rotating}
\usepackage{bbm}
\usepackage{amsmath}
\usepackage{color, colortbl}
\usepackage{soul}
\usepackage{subcaption}
\usepackage{forloop}
\usepackage{amssymb}

\usepackage{pifont}
\usepackage{array}
\usepackage{enumitem}

\usepackage{algpseudocode}
\usepackage[linesnumbered,ruled,vlined]{algorithm2e}

\usepackage{cuted}
\usepackage{capt-of}

\def\paperID{8018} 
\def\confName{ICCV}
\def\confYear{2025}

\def\Titlename{Latest Object Memory Management \\ for Temporally Consistent Video Instance Segmentation}
\def\titlename{Latest Object Memory Management (LOMM)}
\def\titlenameabb{LOMM}
\def\memory{Latest Object Memory}
\def\memoryabb{LOM}
\def\matching{Decoupled Object Association}
\def\matchingabb{DOA}

\newcommand\blfootnote[1]{%
  \begingroup
  \renewcommand\thefootnote{}\footnote{#1}%
  \addtocounter{footnote}{-1}%
  \endgroup
}

\title{\titlenameabb: \Titlename}

\author{%
Seunghun Lee$^{1, \ast}$ \quad
Jiwan Seo$^{1}$ \quad
Minwoo Choi$^{1}$ \quad
Kiljoon Han$^{1}$ \\
Jahoon Jeong$^{1}$ \quad
Zane Durante$^{2}$ \quad
Ehsan Adeli$^{2, \dagger}$ \quad
Sang Hyun Park$^{1}$ \quad
Sunghoon Im$^{1, \dagger}$ \\ \\
{$^{1}$DGIST, Daegu, Republic of Korea} \quad
{$^{2}$Stanford University, Stanford, CA, USA} \\
}

\begin{document}
\maketitle

\blfootnote{$^\ast$ This work was done while visiting Stanford University.}
\blfootnote{$^\dagger$ Corresponding author.}

\vspace{-5mm}

\begin{abstract}
In this paper, we introduce Latest Object Memory (LOM), a system for robustly tracking and continuously updating the latest states of objects by explicitly modeling their presence across video frames. LOM enables consistent tracking and accurate identity management across frames, enhancing both performance and reliability through the video segmentation process. Building upon LOM, we present Latest Object Memory Management (LOMM) for temporally consistent video instance segmentation, significantly improving long-term instance tracking.
This enables consistent tracking and accurate identity management across frames, enhancing both performance and reliability through the video segmentation process.
Moreover, we introduce \matching~(\matchingabb), a strategy that separately handles newly appearing and already existing objects.  By leveraging our memory system, DOA accurately assigns object indices, improving matching accuracy and ensuring stable identity consistency, even in dynamic scenes where objects frequently appear and disappear.
Extensive experiments and ablation studies demonstrate the superiority of our method over traditional approaches, setting a new state-of-the-art in video instance segmentation. Notably, our \titlenameabb~achieves an AP score of 54.0 on YouTube-VIS 2022, a dataset known for its challenging long videos. Project page: \href{https://seung-hun-lee.github.io/projects/LOMM/}{this https URL}  
\end{abstract}


\section{Introduction}

Video instance segmentation (VIS) is a complex task that requires segmenting, classifying, and tracking objects across video frames \citep{yang2019video}. Recent breakthroughs in this field have been significantly propelled by the adoption of query-based segmentation networks \citep{cheng2021per, cheng2022masked}, which have notably enhanced the precision of instance segmentation. These networks utilize object queries to extract distinctive features for each object within a frame and then group pixels on the image feature map to delineate object regions. Building on the capabilities of these architectures, the latest VIS approaches \citep{huang2022minvis, heo2022vita, wu2022defense, heo2023generalized, ying2023ctvis, DVIS, li2023tcovis, kim2024offline} focus on developing sophisticated methods for instance association, aiming to improve the accuracy and efficiency of tracking objects across a video.

\begin{figure*}[t]
    \centering
    \includegraphics[width=0.82\linewidth]{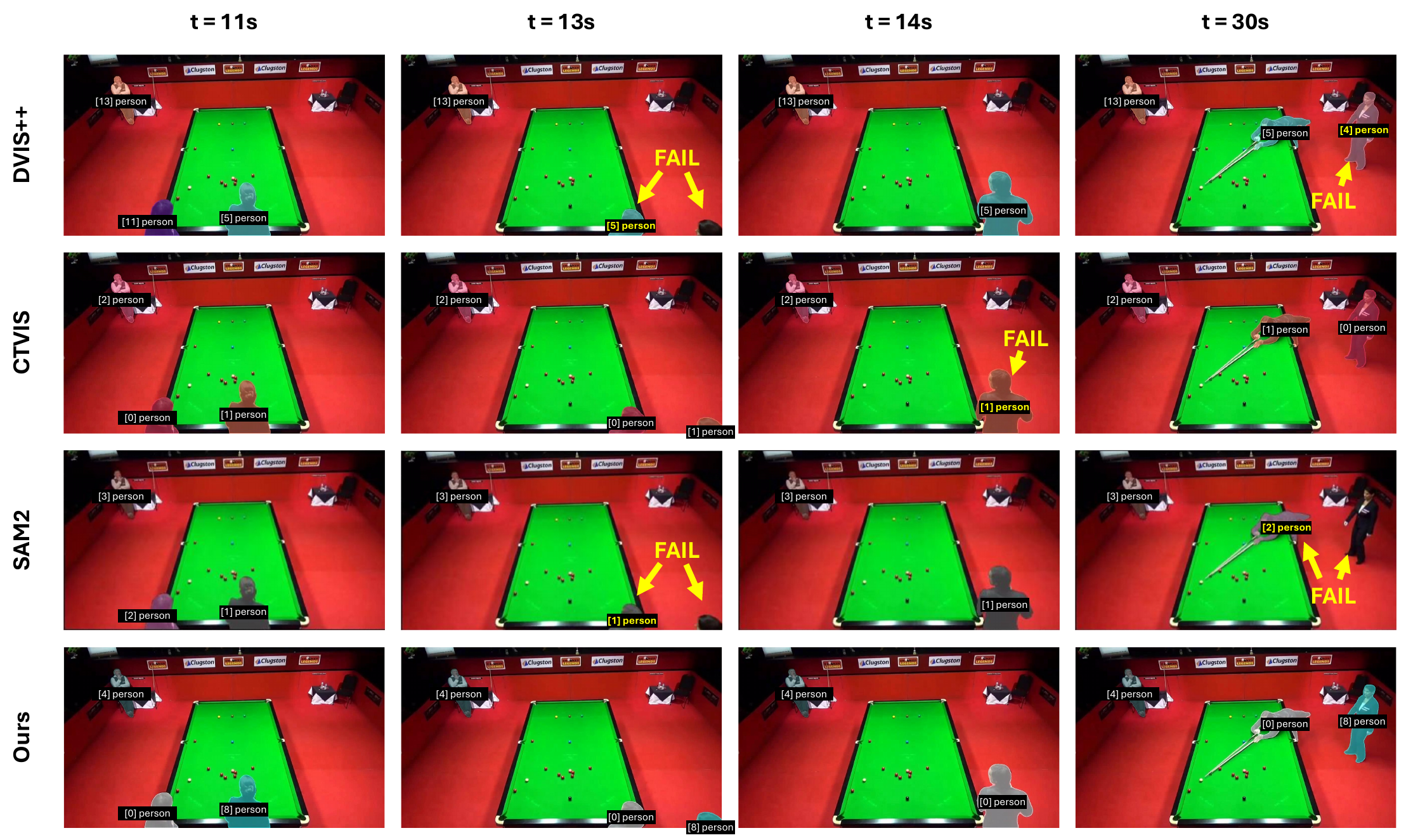}
    \vspace{-3mm}
    \caption{ 
    \textbf{Qualitative comparisons on a challenging video from YouTube-VIS 2022.} \textbf{\textcolor{yellow!85!black}{Yellow text}} highlights objects that failed to be tracked.
        (First row) DVIS++ \cite{zhang2023dvis++} often fails to track objects that reappear after a temporal gap, as it relies solely on the previous frame.
        (Second row) CTVIS \cite{ying2023ctvis} employs a memory mechanism that aggregates object features across frames based on similarity weights, potentially leading to ID switching.
        (Third row) SAM2 \cite{ravi2024sam} stores spatial feature maps alongside object features; however, it fails to effectively associate this information, leading to failures in challenging scenarios.
        (Fourth row) Our approach continuously updates and matches the latest state of each object, enabling robust and consistent tracking.}
    \label{fig:motivation}
    \vspace{-5mm}
\end{figure*}

To enhance object association between frames, transformer-based trackers \citep{DVIS, zhang2023dvis++} directly align predictions using cross-frame attention, dynamically reconstructing object representations from prior frames. While effective for short-term tracking, these methods heavily rely on previous frame data, which can hinder long-term consistency, as shown in the first row of \figref{fig:motivation}. To mitigate this issue, some approaches incorporate a memory system, updating object representations based on similarity \citep{wu2022defense, ying2023ctvis, kim2023visage} or momentum mechanisms \citep{gao2023memotr}. However, as shown in the second row of \figref{fig:motivation}, these methods often struggle due to mixed representations, causing ambiguity in memory-object matching and leading to tracking failures. Even SAM2 \cite{ravi2024sam}, which stores not only object features but also spatial feature maps in memory, fails to maintain consistency in long videos, as demonstrated in the third row of \figref{fig:motivation}. These limitations highlight the need for a more robust approach that consistently updates and associates object representations over long videos. (We refer the reader to the \textbf{Supplementary Material} for deeper understanding.)

To tackle the challenges inherent in VIS, we introduce \textit{\titlename} designed for temporally consistent object tracking. As its core, \textit{\memory~(\memoryabb)} maintains and updates the latest state of each object, using the most recent object state as the key cue for tracking. By explicitly modeling object presence in each frame, our method efficiently handles dynamic scene where  objects frequently appear and disappear.
This memory-driven approach ensures stable identity management across frames, significantly improving VIS accuracy and reliability.
We also introduce an \textit{\matching~(\matchingabb)}, a method that leverages our memory system for precise object matching. This strategy separates the object association process into two stages: (1) tracking existing objects using stored memory states to maintain consistent identities and (2) assigning new IDs to newly appeared objects using our Occupancy-guided Hungarian Matching (OHM). By managing unoccupied indices, our method preserves existing object identities while seamlessly integrating new one.

Our approach improves upon traditional cross-frame attention mechanisms \cite{DVIS}, which often misassign new objects to background indices, leading to  learning conflicts. Instead of relying solely on object features from the previous frame as anchor queries, we incorporate DOA to determine object IDs in the current frame, enabling construction of adaptive anchor queries for more robust matching. By separately handling existing and new objects, our decoupled strategy ensures precise and consistent object alignment across frames, overcoming key limitations of prior methods. Extensive experiments and ablation study demonstrate that the effectiveness of our method, achieving superior performance over existing techniques.

Our contributions can be summarized as follows:
\begin{itemize}[leftmargin=15pt]
    \item[1.] We present a new memory-based video instance segmentation, called \textit{\titlename}, which dynamically manages the latest object states for temporally consistent object matching.
    \item[2.] We introduce \textit{\memory~(\memoryabb)}, which accurately updates the most recent object features, crucial for precise object matching.
    \item[3.] We propose \textit{\matching~(\matchingabb)}, which matches existing objects in memory and assigns new IDs to newly detected ones.
    \item[4.] LOMM excels in challenging long videos, achieving state-of-the-art performance in VIS benchmarks.
\end{itemize}



\section{Related Works}

\textbf{Video Instance Segmentation.}
VIS primarily learns frame-to-frame feature associations using instance segmentation architectures. The seminal work, MaskTrack R-CNN \citep{yang2019video}, add a tracking head to Mask R-CNN \citep{he2017mask}, enhancing instance association. This is advanced by SipMask \citep{cao2020sipmask} and CrossVIS \citep{yang2021crossover}, which improves temporal associations with cross-frame learning. Additionally, IDOL \citep{wu2022defense} incorporates contrastive learning into a query-based architecture \citep{zhu2020deformable}, boosting online method performance. Beyond online methods, VisTR \citep{wang2021end} apply DETR \citep{carion2020end} for clip-level instance predictions, constrained by dense self-attention. Efficiency improvements are further pursued with IFC \citep{hwang2021video}, introducing a transformer with separate spatial and temporal attentions, and TeViT \citep{yang2022temporally} and Seqformer \citep{wu2022seqformer}, which adapt vision transformer backbones to enhance temporal associations and video-level instance predictions, respectively.
Recently, query-based segmentation networks have become fundamental to contemporary VIS approaches, prominently featuring Mask2Former \citep{cheng2022masked} as a key underlying technology. MinVIS \citep{huang2022minvis} simplifies tracking by using post-processing based on cosine similarity between object features. VITA \citep{heo2022vita} enhances this by temporally associating frame-level queries to identify instance prototypes within a video. GenVIS \citep{heo2023generalized} creates a tracking network that operates at the sub-clip level. CTVIS \citep{ying2023ctvis} employs contrastive learning across an expanded frame set to achieve detailed frame associations. DVIS \citep{DVIS} introduces a decoupled architecture that segments the processes into distinct tasks of segmentation, tracking, and refinement.

\textbf{Object Tracking with Memory.}
Memory-based methods have shown significant advancements in video analysis, particularly in tasks requiring sustained long-term consistency, such as video object segmentation \citep{tokmakov2017learning, xu2018youtube, duarte2019capsulevos, ventura2019rvos, huang2020fast, zhang2020transductive, cheng2022xmem}, video instance segmentation \citep{yang2019video, wu2022defense, heo2023generalized, ying2023ctvis, lee2024context}, and video object tracking \citep{yang2018learning, fu2021stmtrack, yan2021learning, cai2022memot, meinhardt2022trackformer, zhao2023object, gao2023memotr}. 
Some research has successfully utilized external memory in multi-object tracking scenarios. MaskTrack R-CNN \citep{yang2019video} employs an external memory to store predicted instance representations, updating them with new data from the latest frame through a straightforward replacement rule. Building on this, with the introduction of DETR \citep{carion2020end}, Meinhardt et al. \citep{meinhardt2022trackformer} developed TrackFormer, which innovates a \textit{tracking-by-attention} approach. This method uses object tokens to maintain temporal instance memories, enhancing tracking continuity. Recent advancements have further improved long-term consistency by either stacking these tokens in a memory buffer \citep{cai2022memot} or applying momentum to update the tokens \citep{gao2023memotr}. Despite these advancements, current techniques often fall short in addressing challenges related to the initial appearance and eventual disappearance of objects, especially due to occlusion. Even unified models \cite{yan2022towards, wang2023omnitracker, yan2023universal, wu2024general} struggle to handle such cases effectively. This gap indicates significant potential for further improvements in memory-based tracking methods.


\section{Preliminary}
\label{sec:tracker}

\textbf{Transformer-based Tracker.}
Video instance segmentation (VIS) involves segmenting and tracking objects consistently across video frames. To address this challenge, recent studies \citep{huang2022minvis, heo2022vita, heo2023generalized, ying2023ctvis, li2023tcovis, DVIS, zhang2023dvis++} have adopted query-based segmentation network like Mask2Former \citep{cheng2022masked}. The segmentation network $\mathcal{S}$ generates object representations $\tilde{Q}_{t} \in \mathbb{R}^{N \times C}$, categorical probabilities $P_{t} \in \mathbb{R}^{N \times (K+1)}$, and segmentation masks $M_{t} \in \mathbb{R}^{N \times H \times W}$ for each video frame $\{I_{t}\}^{T}_{t=1}$ as follows:
\begin{equation}
    \begin{gathered}
        \left[ \tilde{Q}_{t}, P_{t}, M_{t} \right] = \mathcal{S} \left( I_t \right),~~\forall t =  \{1, \ldots ,T \},
    \end{gathered}
\end{equation}
where $N$ is a sufficiently large number to detect objects in an image, while $H$, $W$, and $C$ denote the height, width of predicted mask and the channel dimensions of the object representations, respectively.
The categorical prediction head in the segmentation network classifies objects into $K$ categories or as no-object $\varnothing$.
The class label $c_t \in \mathbb{R}^{N}$ is determined by applying the argmax operation to the probability matrix $P_t$ along the $(K+1)$ dimension.

\begin{figure}
    \centering
    \includegraphics[width=1\linewidth]{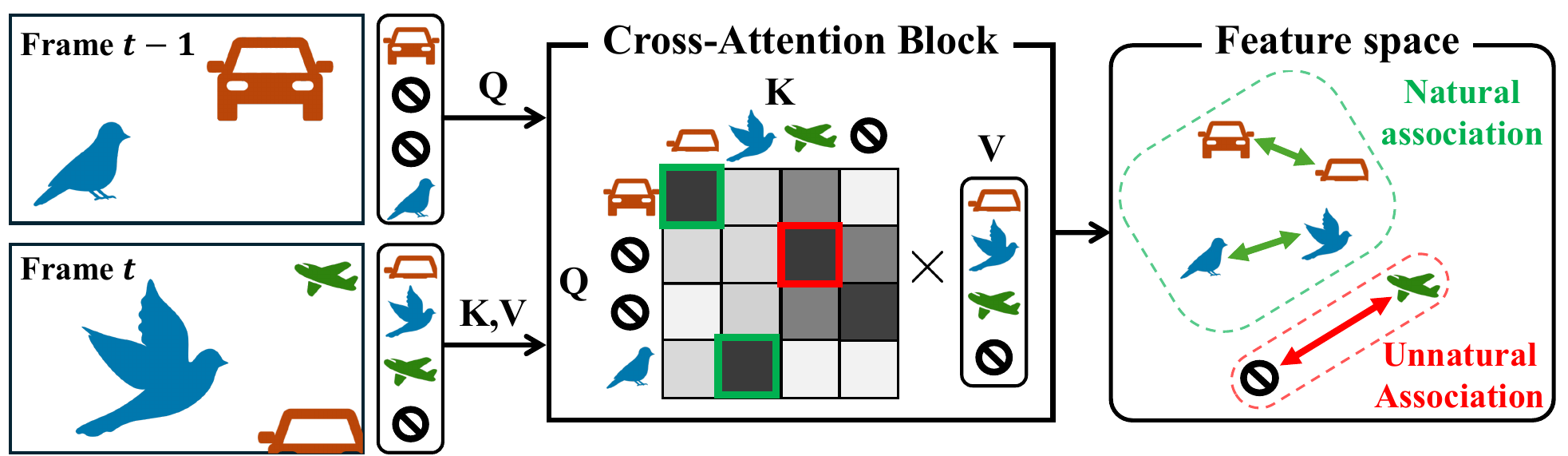}
    \caption{\textbf{Limitations of existing trackers.} When a new object appears, there is no reference for assigning its ID, so it must be mapped to an ID previously used for the background. This results in an unnatural learning process where foreground and background features become overly similar, leading to suboptimal performance.}
    \label{fig:cross-attention}
    \vspace{-5mm}
\end{figure}

The principal challenge in VIS lies in maintaining consistent object representations across frames, ensuring that the aligned sequences of predicted objects ${Q}_{t} \in \mathbb{R}^{N \times C}$ consistently correspond to the same physical entities throughout the video, as follows:
\begin{equation}
\label{eq:tracker}
    {Q}_{t} = \mathcal{T} \left( {Q}_{t-1}, \tilde{Q}_{t} \right),~\forall t \in \{1,\ldots, T\},
\end{equation}
where $\mathcal{T}$ is a tracking network consisting of multiple transformer blocks and the initial queries $Q_0$ are initialized as raw features using raw features extracted from the segmentation network, $\tilde{Q}_1$, from the first frame.
Recent state-of-the-art methods \citep{heo2023generalized, DVIS, zhou2024dvis} use a transformer-based tracker that encodes object features from each frame, informed by the sequence from previous frames. Cross-attention layers within these transformers align current frame objects with previous ones based on feature similarity, ensuring accurate and consistent tracking across the video.

\begin{figure*}[t!]
    \centering
    \includegraphics[width=0.9\linewidth]{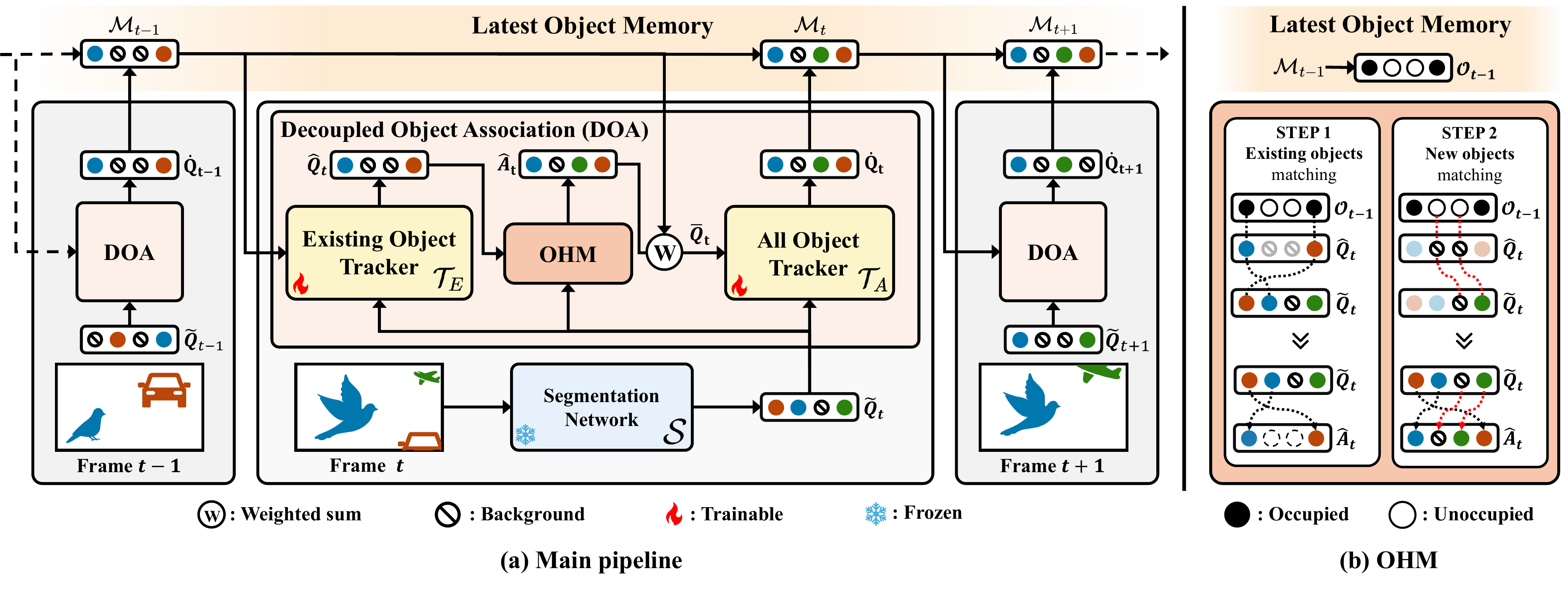}
    \vspace{-4mm}
    \caption{\textbf{Overall pipeline of our \titlenameabb.} (a) For each frame, the existing object tracker $\mathcal{T}_E$ utilizes $\mathcal{M}_{t-1}$ from the \memory~and $\tilde{Q}_{t}$ from the segmentation network $\mathcal{S}$ to predict aligned features $\hat{Q}_{t}$ for matched existing objects. The current frame’s existing objects $\hat{Q}_{t}$ and all objects $\tilde{Q}_{t}$ are matched through Occupancy-guided Hungarian Matching (OHM), ensuring accurate ID assignment for potential new objects and generating aligned object features $\hat{A}_{t}$. For robust tracking, an adaptive query $\bar{Q}_{t}$ is created by blending past information $\mathcal{M}_{t-1}$ with the current frame's aligned objects $\hat{A}_{t}$. This adaptive query $\bar{Q}_{t}$ and the current frame’s objects $\tilde{Q}_{t}$ are processed by $\mathcal{T}_{A}$ to obtain the final tracking result $\dot{Q}_{t}$. (b) OHM assigns newly appeared objects to unoccupied slots using the occupancy prior $\mathcal{O}_{t-1}$. To ensure precise matching, OHM first assigns IDs to existing objects at occupied indices. Then, it allocates IDs to newly appeared objects at unoccupied indices, producing the aligned object features $\hat{A}_{t}$ as output.}
    \label{fig:main}
    \vspace{-3mm}
\end{figure*}

\textbf{Discussion.}
The tracking network described in \equref{eq:tracker} encounters two primary challenges. First, it relies on maintaining consistent object indices once they are assigned, but this consistency is not guaranteed. Since object association depends on matching objects across consecutive frames, objects that temporarily disappear and later reappear are at risk of being assigned incorrect indices. Although exisiting apporaches, such as updating long-term memory with momentum \citep{gao2023memotr} or enhancing object-wise similarity \citep{wu2022defense, ying2023ctvis, kim2023visage} aim to mitigate this, they often fall short of reliably preserving object identities over time.

The second challenge relates to the handling of newly appearing objects. The cross-attention mechanism reconstructs objects by comparing them to those already identified. However, new objects lack any prior reference, meaning they should be assigned to previously unallocated indices. For this to work, their features would need to resemble background features at first--an unnatural and impractical constraint, as shown in \figref{fig:cross-attention}. This leads to ambiguous feature learning, unltimately degrading the system's ability to accurately track newly appearing objects.
To address these issues, we introduce an \titlename~described in \secref{sec:o2vis}.

\section{\titlenameabb}
\label{sec:o2vis}
As illustrated in \figref{fig:main}, our model consists of two tracking networks, $\mathcal{T}_E$ for existing objects and $\mathcal{T}_A$ for all objects including potential new objects. These two networks align the object features $\tilde{Q}$ produced by the pretrained segmentation network $\mathcal{S}$. 
Our approach incorporates two principal components to enhance tracking precision and consistency: a \memory~in \secref{sec:memory} and a \matching~strategy in \secref{sec:reconstruction}. These sections detail how they contribute to maintaining object continuity and seamlessly integrating potential new objects into the scene.

\subsection{\memory}
\label{sec:memory}
To enhance tracking in video instance segmentation (VIS), we first introduce \memory, a crucial component of our \titlenameabb~framework.
This memory system, denoted as $\mathcal{M}_{t}$, continuously updates object features online at each time step $t$. 
Specifically, it accumulates the aligned object features $\dot{Q}_t \in \mathbb{R}^{N\times C}$ over time as follows:
\begin{align}
\begin{split}
    \mathcal{M}_{t} = (1 - p_{t}) \mathcal{M}_{t-1} + p_{t} \dot{Q}_{t},
    ~\forall t = \{1,\ldots,T\},
\end{split}
\end{align}
where $p_t$ is the foreground probability for each object index. This probability is computed by summing the categorical probabilities $\{P_t^k\}_{k=1}^K$ across all object classes, excluding the no-object probability $P_t^{(K+1)}$. 
Our memory is initialized at $\mathcal{M}_0$ with $\dot{Q}_1$ and the aligned queries $\dot{Q}_t$ are defined in \equref{eq:anc_generation}.
This dynamic update process ensures consistent and robust maintenance of object information throughout the video sequence.
By adaptively incorporating changes from the current frame, the memory effectively tracks and updates object representations as they appear, move, and potentially disappear across the video. Unlike traditional memory update methods that might prioritize background information, our method focuses on foreground objects. This prioritization is crucial in dynamic video environments where objects frequently appear and disappear, ensuring that our system maintains accurate and reliable object information over time.

\subsection{\matching}
\label{sec:reconstruction}

Based on our observation described in \secref{sec:tracker}, we design a \matching~strategy that consists of the following steps: (1) existing object tracking, (2) index assignment to new objects, and (3) object alignment with an adaptive anchor query.
To avoid unnatural associations that can arise from directly tracking all objects observed in the instance queries $\tilde{Q}_{t}$, we initially focus on matching all instances to the existing objects recorded in the memory $\mathcal{M}_{t-1}$ as follows:
\begin{equation}
    \hat{Q}_{t} = \mathcal{T}_{E} \left( \mathcal{M}_{t-1}, \tilde{Q}_{t} \right),
\end{equation}
where $\hat{Q}_{t}$ represents aligned features of matched existing objects, and $\mathcal{T}_{E}$ denotes the existing object tracker. This mechanism specifically focuses on identifying existing objects in the current frame, thereby minimizing ambiguity in the object association process.
It is important to note that $\hat{Q}_{t}$ excludes any objects that have newly appeared at time $t$. This ensures that the system accurately tracks and updates only those objects that persist across frames, without being confounded by newly introduced elements.

When new objects appear, they should be assigned previously unallocated indices for accurate tracking. 
To manage the allocation state of each object, we define occupancy statuses $\mathcal{O}$ as follows:
\begin{align}
\begin{split}
    \mathcal{O}_{t} = \mathcal{O}_{t-1} \lor O_{t}, ~~
    O_{t}^n =
    \begin{cases}
        0 & \text{if} ~ c_t^n = \varnothing, \\
        1 & \text{otherwise}\\
    \end{cases}, 
\end{split}
\end{align}
where $c_t^n$ represents the predicted class, and $\varnothing$ denotes the no-object class. $\mathcal{O}_0$ is initialized as a zero vector of length $N$, and each slot $n$ is marked as occupied once a foreground object has been assigned to it at least once.

In our framework, the queries from segmentation network $\tilde{Q}_{t}$ encompasses both existing and newly appeared objects, whereas the queries $\hat{Q}_{t}$ contains only the existing objects at occupied indices. By matching these two object features, appropriate indices for the new objects can be identified. To facilitate this, we introduce an occupancy-guided Hungarian matching mechanism. This ensures that newly appeared objects are not erroneously matched to indices that are already occupied $\mathcal{O}_{t-1}$ from the previous time $(t-1)$.
The matching starts by aligning objects corresponding to the occupied indices in $\hat{Q}_{t}$, followed by the remaining objects.

To ensure robust object association, we combine the aligned object queries $\hat{A}_t$ from Alg. \ref{ohm} with the latest object queries stored in $\mathcal{M}_{t-1}$ to construct an adaptive anchor query for $\mathcal{T}_A$.
Then, we reconstruct the current object features $\hat{Q}_{t}$ using the adaptive anchor query through the alignment network $\mathcal{T}_{A}$ as follows:
\begin{gather}
\begin{split}
    \dot{Q}_{t} & = \mathcal{T}_{A} \left( \bar{Q} , \tilde{Q}_{t} \right),  ~~ \bar{Q}_{t} = \text{Adp}(\hat{A}_t,\mathcal{M}_{t-1})
\label{eq:anc_generation}
\end{split}
\end{gather}
where $\text{Adp}(A,B)=\text{sim} (A,B) \cdot A + \big(1-\text{sim} (A,B) \big)\cdot B$ and $\text{sim}(\cdot,\cdot)$ measures cosine similarity.

\setlength{\textfloatsep}{8pt}
\begin{algorithm}[t]
\DontPrintSemicolon
\textbf{Input}: Existing objects $\hat{Q}_{t} \in \mathbb{R}^{N \times C}$, Current objects $\tilde{Q}_{t} \in \mathbb{R}^{N \times C}$, Occupancy statuses $\mathcal{O}_{t-1} \in \mathbb{B}^N$ \\
\textbf{Output}: Aligned object queries $\hat{A}_{t} \in \mathbb{R}^{N \times C}$\\
    
\textbf{Initialize}: $\hat{A}_{t} \leftarrow$ empty matrix of size $N \times C$

\textbf{Get index sets}:
\quad $ \mathcal{I}_{\text{o}} \leftarrow \{i \mid \mathcal{O}_{t-1}^{i} = 1, \, i \in \{1, \ldots, N\}\}$
\quad $ \mathcal{I}_{\text{u}} \leftarrow \{i \mid \mathcal{O}_{t-1}^{i} = 0, \, i \in \{1, \ldots, N\}\}$

\textbf{Get object features for each index set}:
\quad $ f_{\text{o}} \leftarrow \hat{Q}_{t}[\mathcal{I}_{\text{o}}]$,
\quad $ f_{\text{u}} \leftarrow \hat{Q}_{t}[\mathcal{I}_{\text{u}}]$

\textbf{Find index set for occupied objects}:
\quad $\mathcal{J}_{\text{o}} \leftarrow \text{Hungarian}(f_{\text{o}}, \tilde{Q}_{t})$

\textbf{Get unmatched features in $\tilde{Q}_{t}$}:
\quad $\tilde{f}_{\text{u}} \leftarrow \tilde{Q}_{t} \setminus \{\tilde{Q}_{t}[\mathcal{J}_{\text{o}}]\}$

\textbf{Find index set for unoccupied objects}:
\quad $\mathcal{J}_{\text{u}} \leftarrow \text{Hungarian}(f_{\text{u}}, \tilde{f}_{\text{u}})$

\textbf{Assign aligned object query}:

\quad $\hat{A}_{t}[\mathcal{I}_{\text{o}}] \leftarrow \tilde{Q}_{t}[\mathcal{J}_{\text{o}}]$,
\quad $\hat{A}_{t}[\mathcal{I}_{\text{u}}] \leftarrow \tilde{Q}_{t}[\mathcal{J}_{\text{u}}]$

\textbf{Return} $\hat{A}_{t}$
\caption{\small{Occupancy-guided Hungarian Matching}}
\label{ohm}
\end{algorithm}

\subsection{Training}
\textbf{Early training.} 
In the initial stages of training, the quality of representations, such as $\hat{Q}_{t}$, can be poor, making effective object matching as outlined in Alg. \ref{ohm} challenging. To improve early training outcomes and provide a more stable foundation for learning, we adopt the following approach:

\begin{gather}
\begin{split}
\bar{Q}_{t} & = \text{Adp} (\tilde{Q}^{*}_t, \mathcal{M}_{t-1}), \\ \tilde{Q}^{*}_t & = \text{Hungarian} \left( \tilde{Q}^{*}_{t-1}, \tilde{Q}_{t} \right),
\end{split}
\end{gather}
where ``Hungarian" refers to the Hungarian matching algorithm \citep{kuhn1955hungarian}, used to improve the initial alignment of object representations between frames. The initial queries, $\tilde{Q}^*_0$, are initialized as raw features extracted from the segmentation network, $\tilde{Q}_1$, in the first frame.

The initial outputs from the tracking networks \( \mathcal{T}_E \) and \( \mathcal{T}_A \) often exhibit considerable noise, which can impede the accuracy of the tracking process. To mitigate this, we adopt the approach described in \citep{DVIS} for ground truth assignment. Specifically, predictions \( \hat{y} \) from \( \tilde{Q}^*_t \) are used for assigning ground truth via Hungarian matching. This method is strategically implemented during the first half of the total training iterations. This phased application allows the model to adapt incrementally to the task's complexity, enhancing the quality of the training representations as the process evolves. Such a staged training approach not only stabilizes the learning curve but also significantly improves alignment and tracking accuracy over time (see \cref{tab:ablation}-(f)).


\begin{table*}[t!]
\centering
\caption{Comparison on YouTube-VIS validation sets. $*$ denotes offline methods. We employ a temporal refiner \cite{DVIS} for offline model.}
\vspace{-3mm}
\begin{adjustbox}{max width = 0.95\textwidth}
\begin{tabular}{lc|ccccc|ccccc|ccccc}
\specialrule{1.5pt}{1pt}{1pt}
\multicolumn{1}{c}{\multirow{2}{*}{Method}}  & \multirow{2}{*}{Backbone}  & \multicolumn{5}{c|}{YouTube-VIS 2019}  & \multicolumn{5}{c|}{YouTube-VIS 2021} & \multicolumn{5}{c}{YouTube-VIS 2022}  \\
& & \multicolumn{1}{l}{AP} & \multicolumn{1}{l}{AP$_{50}$} & \multicolumn{1}{l}{AP$_{75}$} & \multicolumn{1}{l}{AR$_{1}$} & \multicolumn{1}{l|}{AR$_{10}$} & \multicolumn{1}{l}{AP} & \multicolumn{1}{l}{AP$_{50}$} & \multicolumn{1}{l}{AP$_{75}$} & \multicolumn{1}{l}{AR$_{1}$} & \multicolumn{1}{l|}{AR$_{10}$} & \multicolumn{1}{l}{AP} & \multicolumn{1}{l}{AP$_{50}$} & \multicolumn{1}{l}{AP$_{75}$} & \multicolumn{1}{l}{AR$_{1}$} & \multicolumn{1}{l}{AR$_{10}$} \\ \midrule
MinVIS \citep{huang2022minvis}& R50  & 47.4  & 49.0  & 52.1  & 45.7  & 55.7& 44.2  & 66.0  & 48.1  & 39.2  & 51.7 & 23.3  & 47.9  & 19.3  & 20.2  & 28.0 \\
VITA$^*$ \citep{heo2022vita}  & R50 & 49.8 & 72.6 & 54.5 & 49.4 & 61.0 & 45.7 & 67.4 & 49.5 & 40.9 & 53.6  & 32.6  & 53.9  & 39.3  & 30.3  & 42.6 \\
GenVIS \citep{heo2023generalized} & R50  & 50.0 & 71.5 & 54.6 & 49.5 & 59.7 & 47.1 & 67.5 & 51.5 & 41.6 & 54.7& 37.5  & \underline{61.6}  & 41.5  & 32.6  & 42.2 \\
DVIS \citep{DVIS}  & R50  & 51.2 & 73.8 & 57.1 & 47.2 & 59.3 & 46.4 & 68.4 & 49.6 & 39.7 & 53.5 & 31.6  & 52.5  & 37.0  & 30.1  & 36.3 \\
TCOVIS \citep{li2023tcovis} & R50  & 52.3 & 73.5 & 57.6 & 49.8 & 60.2 & 49.5 & 71.2 & 53.8 & 41.3 & 55.9& \underline{38.6}  & 59.4  & \underline{41.6}  & \underline{32.8}  & \underline{46.7} \\
DVIS-DAQ \citep{zhou2024dvis} & R50 & 55.2 & 78.7 & \textbf{61.9} & 50.6 & \textbf{63.7} & \underline{50.4} & \underline{72.4} & \underline{55.0} & 41.8 & 57.6  & 34.6  & -  & 35.5  & -  & 41.1 \\
DVIS++ \citep{zhang2023dvis++} & R50  & \underline{55.5} & \textbf{80.2} & 60.1 & \underline{51.1} & 62.6 & 50.0 & 72.2 & 54.5 & \underline{42.8} & \textbf{58.4}& 37.2  & 57.4  & 40.7  & 31.8  & 44.6 \\
Ours  & R50    & \textbf{55.7} & \underline{79.8} & \underline{61.4} & \textbf{51.3} & \underline{62.7}  & \textbf{50.7} & \textbf{72.9} & \textbf{56.9} & \textbf{43.7} & \textbf{58.4} & \textbf{41.1} & \textbf{62.4} & \textbf{46.2} & \textbf{35.8} & \textbf{47.5} \\ \midrule
MinVIS \citep{huang2022minvis} & Swin-L& 61.6 & 83.3 & 68.6 & 54.8 & 66.6 & 55.3 & 76.6 & 62.0 & 45.9 & 60.8 & 33.1  & 54.8  & 33.7  & 29.5  & 36.6\\
VITA$^*$ \citep{heo2022vita} & Swin-L& 63.0 & 86.9 & 67.9 & 56.3 & 68.1 & 57.5 & 80.6 & 61.0 & 47.7 & 62.6 & 41.1  & 63.0  & 44.0  & 39.3  & 44.3\\
DVIS \citep{DVIS} & Swin-L&63.9 & 87.2 & 70.4 & 56.2 & 69.0 & 58.7 & 80.4 & 66.6 & 47.5 & 64.6& 39.9  & 58.2  & 42.6  & 33.5  & 44.9\\
GenVIS \citep{heo2023generalized} & Swin-L& 64.0 & 84.9 & 68.3 & 56.1 & 69.4 & 59.6 & 80.9 & 65.8 & 48.7 & 65.0 & 45.1  & 69.1  & 47.3  & 39.8  & 48.5 \\
TCOVIS \citep{li2023tcovis} & Swin-L& 64.1 & 86.6 & 69.5 & 55.8 & 69.0 & 61.3 & 82.9 & 68.0 & 48.6 & 65.1 & \underline{51.0}  & 73.0  & \underline{53.5}  & \underline{41.7}  & \underline{56.5} \\
DVIS++ \citep{zhang2023dvis++} & ViT-L & 67.7 & 88.8 & 75.3 & 57.9 & \underline{73.7} & 62.3 & 82.7 & 70.2 & 49.5 & 68.0 & 37.5  & 53.7  & 39.4  & 32.4  & 43.5 \\
DVIS-DAQ \citep{zhou2024dvis} & ViT-L & 68.3 & 88.5& 76.1& 58.0& 73.5 & 62.4& 83.6& 70.8& 49.1& 68.0& 42.0  & -  & 43.0  & -  & 48.4\\
DVIS++$^*$ \citep{zhang2023dvis++} & ViT-L & 68.3& \underline{90.3}& 76.1& 57.8& 73.4 &63.9 &\underline{86.7} &71.5 &48.8 & \underline{69.5}& 50.9  & \underline{75.7}  & 52.8  & 40.6  & 55.8 \\
Ours  & ViT-L   & \underline{69.1} & 89.3 & \underline{76.5} & \underline{58.1} & 73.5 & \underline{65.0} & 86.0 & \underline{72.7} & \underline{49.6} & 69.1 & 48.2 & 70.5 & 53.2 & 40.7 & 52.6\\ 
Ours$^*$  & ViT-L & \textbf{70.1} & \textbf{90.7} & \textbf{77.7} & \textbf{58.5} & \textbf{74.8}  & \textbf{66.2} & \textbf{88.6} & \textbf{74.9} & \textbf{49.9} & \textbf{70.6} & \textbf{54.0} & \textbf{77.1} & \textbf{57.9} & \textbf{43.2} & \textbf{58.8}\\
\specialrule{1.5pt}{1pt}{1pt}
\label{tab:ytvis}
\end{tabular}
\end{adjustbox}
\vspace{-6mm}
\end{table*}

\textbf{Training loss.}
In our approach, we utilize a comprehensive loss function aligned with those \citep{cheng2021mask2former, li2023tcovis, DVIS}.
This function incorporates categorical cross-entropy, binary cross-entropy, and dice losses, which are pivotal for effectively training our model. We specifically focus on optimizing the networks \( \mathcal{T}_E \) and \( \mathcal{T}_A \), while keeping other parameters static. 
We use both the actual ground truth $y_t$ and a modified version, \(\mathring{y}_{t}^{n}\), which adapts the indices by assigning the no-object label \(\varnothing\) and zero mask where new objects appear.
This modification ensures that \(\mathring{y}_{t}\) reflects only predictions for existing objects, thereby enhancing the relevance and accuracy of the training process.
The structure of our primary training loss is computed as follows:
\begin{equation}
\label{eq:inst}
\begin{gathered}
\mathcal{L}_{\text{Track}} = \sum_{t=1}^{T} \sum_{n=1}^{N_{GT}} \left( \mathcal{L} \left( \mathring{y}_{t}^n, \hat{y}_{t}^{\acute{\sigma}(n)} \right) + \mathcal{L} \left( y_{t}^n, \dot{y}_{t}^{\acute{\sigma}(n)}  \right) \right), \\
\acute{\sigma}= \underset{\sigma\in \mathfrak{S}_{N}}{\arg \min } \sum_{n=1}^{N_{GT}} \mathcal{L}_{\text {Match }}\left(y_{f(n)}^n, \dot{y}_{f(n)}^{\sigma(n)}\right),
\end{gathered}
\end{equation}
where \(\dot{y}_{t}\) and \(\hat{y}_{t}\) are the predictions from the feature representations \(\dot{Q}_{t}\) and \(\hat{Q}_{t}\), respectively. 
$\mathfrak{S}_N$ represents a permutation of $N$ elements, and 
$\mathcal{L}_\text{Match}$ denotes a pair-wise matching cost \citep{cheng2021mask2former}.
$f(n)$ is computed specifically for the frame in which each $n$-th object first appears, following the approach used in \citep{DVIS}.
This ensures that our model specifically learns from relevant, existing object features and avoids any confusion from non-object areas or noise.

To ensure that occupied indices in $\tilde{Q}$ consistently represent the same objects across frames and that unoccupied indices reflect changes, we implement a similarity-based loss function as follows:
\begin{equation}
    \mathcal{L}_{\text{Sim}} = \frac{1}{T N_{GT}} \sum_{t=2}^{T} \sum_{n=1}^{N_{GT}} \bigg( \text{sim} \left(\hat{A}_t^{\acute{\sigma}(n)}, \mathcal{M}_{t-1}^{\acute{\sigma}(n)} \right) - O_{t-1}^{\acute{\sigma}(n)} \bigg)^2.
\end{equation}
Our model is jointly trained using an objective function that combines the tracking loss and the similarity loss, with a balance determined by the weight \( \lambda_{\text{Sim}} \):
\begin{equation}
    \mathcal{L}_{\text{Total}} = \mathcal{L}_{\text{Track}} + \lambda_{\text{Sim}} \mathcal{L}_{\text{Sim}}.
\end{equation}

\section{Experiments}

\subsection{Implementation Details}
We evaluate the performance of our \titlenameabb~using standard benchmark datasets: YouTubeVIS datasets (2019, 2021, 2022) \citep{yang2019video} and OVIS \citep{qi2022occluded}.
For our segmentation network, we employ the Mask2Former architecture \citep{cheng2022masked} equipped with three distinct backbone encoders: ResNet-50 \citep{he2016deep}, Swin-L \cite{liu2021swin}, ViT-L and ViT-H \citep{dosovitskiy2021an}. 
All backbones are initialized with parameters pre-trained on COCO \citep{lin2014microsoft}.
Our tracking framework integrates two networks $\mathcal{T}_E$ and $\mathcal{T}_A$, each comprising three transformer blocks and enhanced with a referring cross-attention layer \citep{DVIS} for improved accuracy.
Our tracking networks are trained with all other parameters frozen as previous studies \citep{DVIS, li2023tcovis}. 
We empirically set $\lambda_{\text{Sim}}$ as 1.0. Further details are described in the supplementary materials.


\begin{figure*}
    \centering
    \includegraphics[width=0.90\linewidth]{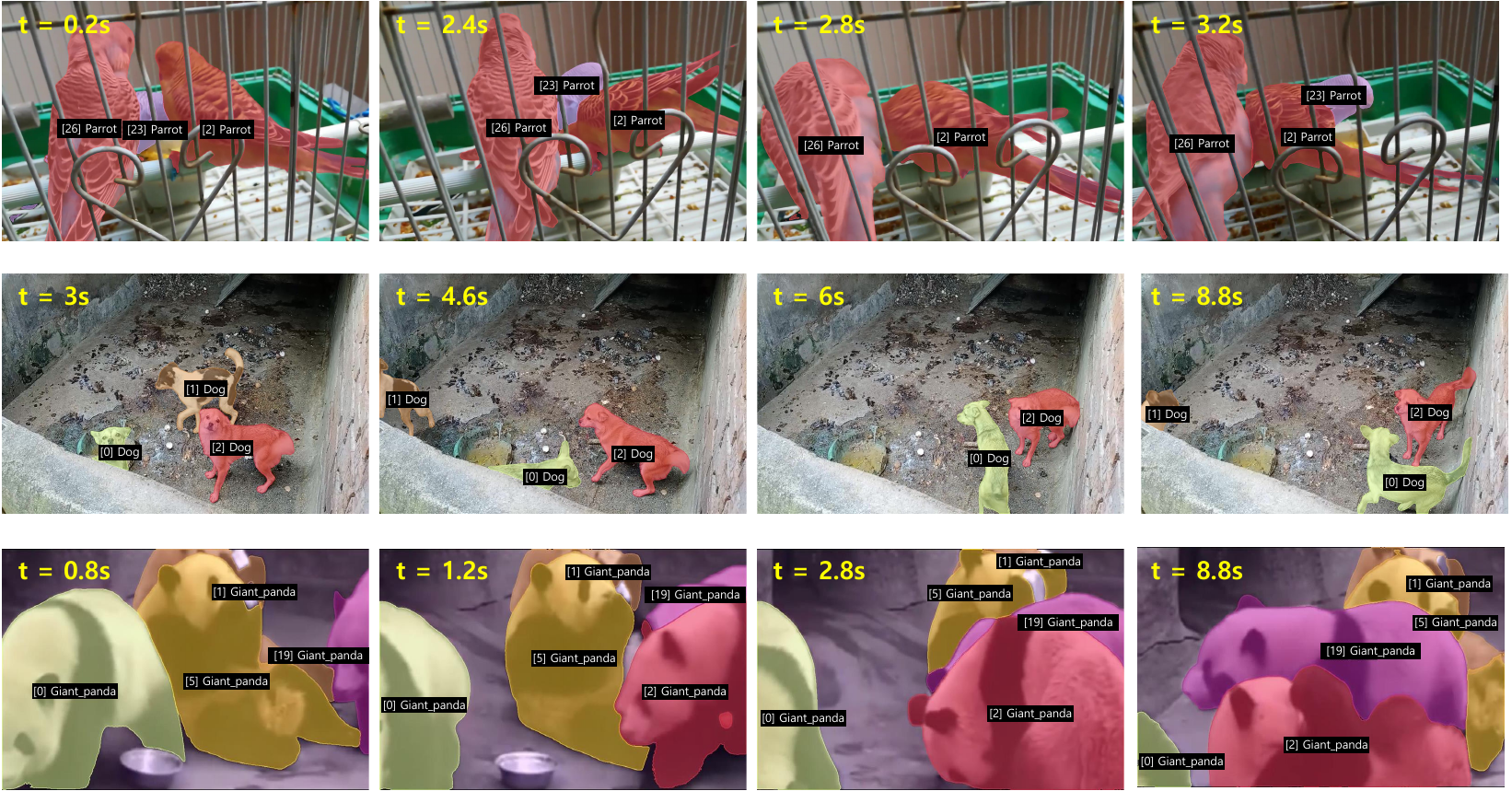}
    \caption{\textbf{VIS results from our model on challenging videos in OVIS dataset.}
    (First and Second row) In both videos, the objects are successfully tracked without any ID switch even after complete occlusion. (Third row) The model robustly tracks a newly appeared object throughout the entire video, even in a crowded scene.}
    \label{fig:ovis_results}
    \vspace{-3mm}
\end{figure*}

\subsection{Main Results}
Following the standard evaluation metrics, Average Precision (AP) and Average Recall (AR), we benchmark the performance of \titlenameabb~against the current state-of-the-art methods in video instance segmentation.

\begin{table}[t!]
\centering
\caption{Comparisons on OVIS validation sets.}
\vspace{-3mm}
\begin{adjustbox}{max width = 0.45\textwidth}
\begin{tabular}{lc|ccccc}
\specialrule{1.5pt}{1pt}{1pt}
Method & Backbone & AP & AP$_{50}$ & AP$_{75}$ & AR$_{1}$ & AR$_{10}$ \\ \specialrule{1.2pt}{1pt}{1pt}
MinVIS\citep{huang2022minvis}      & Swin-L & 39.4 & 61.5 & 41.3 & 18.1 & 43.3 \\
IDOL\citep{wu2022defense}          & Swin-L & 40.0 & 63.1 & 40.5 & 17.6 & 46.4 \\
MDQE\citep{li2023mdqe}             & Swin-L & 41.0 & 67.9 & 42.7 & 18.3 & 45.2 \\
NOVIS\citep{meinhardt2023novis}    & Swin-L & 43.0 & 66.9 & 44.5 & 18.9 & 46.3 \\
GenVIS\citep{heo2023generalized}   & Swin-L & 45.2 & 69.1 & 48.4 & \underline{19.1} & 48.6 \\
DVIS\citep{DVIS}                   & Swin-L & 45.9 & 71.1 & 48.3 & 18.5 & 51.5 \\
TCOVIS\citep{li2023tcovis}         & Swin-L & 46.7 & 70.9 & \underline{49.5} & \underline{19.1} & 50.8 \\
CTVIS\citep{ying2023ctvis}         & Swin-L & \underline{46.9} & \underline{71.5} & 47.5 & \underline{19.1} & \underline{52.1} \\
Ours                               & Swin-L & \textbf{47.8} & \textbf{73.6} & \textbf{51.4} & \textbf{19.2} & \textbf{52.4} \\ \hline
DVIS++\citep{zhang2023dvis++}      & ViT-L & \underline{49.6} & \underline{72.5} & \underline{55.0} & \underline{20.8} & \underline{54.6} \\
Ours                               & ViT-L & \textbf{51.7} & \textbf{73.9} & \textbf{57.5} & \textbf{21.1} & \textbf{56.2} \\ \hline
UNINEXT\citep{yan2023universal}   & ViT-H & \underline{49.0} & \underline{72.5} & \underline{52.2} & -    & -    \\
Ours                               & ViT-H & \textbf{52.9} & \textbf{76.1} & \textbf{55.3} & \textbf{20.1} & \textbf{57.4} \\ \specialrule{1.5pt}{1pt}{1pt}
\end{tabular}
\end{adjustbox}
\label{tab:ovis}
\end{table}


\textbf{Results on Youtube-VIS.}
We compare \titlenameabb~with leading methods on the YouTube-VIS (YTVIS) datasets. The performance metrics, detailed in \tabref{tab:ytvis}, show that \titlenameabb~outperforms state-of-the-art method, DVIS++, by achieving higher AP scores: +1.8 AP on YTVIS19, +2.2 AP on YTVIS21, and +3.1 AP on YTVIS22.
Notably, \titlenameabb~also surpasses the performance of DVIS-DAQ, which addresses newly emerging and disappearing objects, under the same online setting equipped with a ViT-L backbone, achieving margins of +0.8 AP, +2.6 AP, and +6.2 AP on YTVIS19, YTVIS21, and YTVIS22, respectively.
This significant improvement, particularly on longer video sequences, underscores \titlenameabb's ability to maintain long-term consistency effectively. This is largely attributed to its innovative latest object memory mechanism that adapts dynamically to complex video contexts.

\textbf{Results on OVIS.}
Our results on the OVIS benchmark are detailed in \tabref{tab:ovis}, where \titlenameabb~exhibits remarkable superiority over existing models. 
Notably, both with the ViT-L and ViT-H backbone, our model sets a new state-of-the-art by outperforming CTVIS by +0.9 AP, DVIS++ by +2.1 AP, and UNINEXT by +3.9 AP. 
The strong performance across these configurations, particularly in a dataset characterized by frequent occlusions and dynamic object appearances, highlights the efficacy of \titlenameabb's latest-state-aware memory and object association techniques in maintaining accurate and consistent tracking under challenging conditions. As shown in \figref{fig:ovis_results}, our model demonstrates a clear advantage in complex scenes, accurately predicting object trajectories even amid severe occlusions and multiple interacting objects.

\newcommand\x{4pt}

\begin{table*}[t!]
    \centering
    \caption{Ablation studies on each component of \titlenameabb. All experiments report AP scores on the YouTube-VIS 2022 dataset using a ResNet-50 backbone. (a) Effect of \memory~(\memoryabb)~and \matching~(\matchingabb). (b) Performance comparison of different memory systems across baselines. (c) Comparison between prediction from $\hat{A}$ and its refined output $\dot{Q}$ via $\mathcal{T}_{A}$. (d) Performance comparison using $\hat{A}_{t}$ (current frame), $\mathcal{M}_{t-1}$ (past frames), and $\bar{Q}_{t}$ (blended adaptive anchor) as anchors for $\mathcal{T}_{A}$. (e) Comparison between OHM and standard Hungarian matching. (f) Effect of early training.}
    \vspace{-2mm}
    \subfloat[\memoryabb, \matchingabb]{%
        \small{\begin{tabular}{cc|c}
        \toprule
             \memoryabb & \matchingabb & AP \\  
            \hline
               &     & 33.9 \\
              &   \cmark  & 36.2 \\ 
               \cmark&  & 39.2 \\
               \cmark  & \cmark   & \textbf{41.1} \\
            \bottomrule
        \end{tabular}}%
    }\hfill%
    \subfloat[Memory type]{%
        \begin{tabular}{c|cccc}
        \toprule
        \multirow{2}{*}{Baseline} & \multicolumn{4}{c}{Memory type} \\
               & \xmark     & Similarity   & Momentum & LOM  \\ \hline
        \\[-7pt]
        MinVIS \cite{huang2022minvis} & 26.8  & 28.5  & 30.9     & \textbf{34.0} \\[3pt]
        DVIS \cite{DVIS}  & 33.9  & 35.8  & 35.2     & \textbf{39.2} \\[3pt] \bottomrule
        \end{tabular}%
    }\hfill%
    \subfloat[Necessity of $\mathcal{T}_{A}$]{%
        \small{\begin{tabular}{c|c}
        \toprule
        \\[-6pt]
            Prediction from & AP \\[4pt]  
            \hline \\[-8pt]
               $\hat{A}$ & 39.1 \\[4pt]
               $\dot{Q}$  & \textbf{41.1} \\[4pt] 
            \bottomrule
        \end{tabular}}%
    }\hfill%
    \\[10pt]
    \subfloat[Anchor query for $\mathcal{T}_A$]{%
        \normalsize{\begin{tabular}{c|ccc}
        \toprule
            \multirow{2}{*}{ Metric } & \multicolumn{3}{c}{ Anchor query } \\
            & $\hat{A}_t$ & $\mathcal{M}_{t-1}$ & $\bar{Q}_{t}$ \\
            \hline
            AP & 39.3 & 39.2 & \textbf{41.1} \\
            \bottomrule
        \end{tabular}}%
    }\hfill%
    \subfloat[Occupancy-guided Hungarian Matching]{%
        \normalsize{\begin{tabular}{c|cc}
        \toprule
            \multirow{2}{*}{ Metric } & \multicolumn{2}{c}{~~ Occupancy awareness ~~ } \\
            &  \xmark & \cmark \\
            \hline
            AP & 40.3 & \textbf{41.1} \\
            \bottomrule
        \end{tabular}}%
    }\hfill%
    \subfloat[Early training]{%
        \normalsize{\begin{tabular}{c|cc}
        \toprule
            \multirow{2}{*}{ Metric } & \multicolumn{2}{c}{ ~~ Early training ~~ } \\
            & \xmark & \cmark \\
            \hline
            AP & 40.0 & \textbf{41.1} \\
            \bottomrule
            
        \end{tabular}}%
    }%
    \label{tab:ablation}
    \vspace{-3mm}
\end{table*}

\subsection{Ablation Study}

To verify the effectiveness of our method on long videos, we conduct an ablation study on the YouTube-VIS-2022 dataset. We use a ResNet-50 backbone for all experiments and adopt the AP score as the evaluation metric.

\textbf{\memory.}
We demonstrate the effectiveness of our \memory~through the results in \tabref{tab:ablation}-(a) and (b), which show a notable improvement by +7.2 and +5.3 AP compared to the baseline MinVIS \cite{huang2022minvis} and DVIS \cite{DVIS}, respectively.
The conventional memory systems struggle to effectively update recent object information, resulting in suboptimal tracking performance. (Refer to our supplementary materials for a more detailed analysis.) 
This underscores the effectiveness of our approach, confirming that updating memory with foreground probability significantly enhances the model's accuracy and reliability, especially in challenging scenarios where precise object association is essential.

\textbf{\matching.}
\tabref{tab:ablation}-(a), (c), (d), and (e) demonstrates the effectiveness of the \matching, which consists of two trackers, $T_E$ and $T_A$, and Occupancy-guided Hungarian Matching.
Applying DOA to the baseline DVIS \cite{DVIS} results in +3.3 AP improvement, and combining it with LOM achieves a significant synergy, leading to a +7.2 AP gain. The aligned object queries $\hat{A}_t$, obtained by aligning the segmentation network's output through Occupancy-guided Hungarian Matching (OHM), achieve a strong performance of 39.1 AP. Further refining the current frame’s object features $\hat{A}$ using past information $\mathcal{M}_{t-1}$ via $\mathcal{T}_A$ results in the refined object features $\dot{Q}$, achieving an even higher score of 41.1 AP. Using either the current frame’s object $\hat{A}_t$ or the past memory $\mathcal{M}_{t-1}$ alone as the anchor query for $\mathcal{T}_A$ results in suboptimal performance, yielding 39.3 AP and 39.2 AP, respectively. Notably, OHM without occupancy guidance still achieves a competitive 40.3 AP, but incorporating occupancy information enables more robust matching, further boosting performance. These results highlight the effectiveness of our matching strategy in consistently tracking both existing and newly appeared objects.

\textbf{Early training strategy.}
Although \titlenameabb~demonstrates robust performance, the initial stages of training the tracker can be particularly challenging due to the scarcity of learned information. To address this, we adopt a strategy where training initially relies on the outputs from the pre-trained segmentation network $\mathcal{S}$ for the first half of the training iterations. This approach facilitates a more stable and informed beginning to the learning process. The effectiveness of this strategy compared to conventional methods is detailed in \tabref{tab:ablation}-(f), showcasing the advantages of integrating the pre-trained knowledge at the early stages.

\begin{table}[t!]
\centering
\scriptsize
\caption{Trade-off between computational cost and performance.}
\vspace{-3mm}
\resizebox{\linewidth}{!}{%
\begin{tabular}{l|cccc}
\toprule
Tracker    & Params (M) $\downarrow$ & Time (ms) $\downarrow$ & AP (YTVIS22) $\uparrow$ \\ \hline
DVIS \cite{DVIS} & \textbf{9.68} & \textbf{91.26} & 33.9 \\
DVIS++ \cite{zhang2023dvis++} & 26.78  & \underline{97.73} & \underline{37.2} \\
DVIS-DAQ \cite{zhou2024dvis} & 18.62 & 209.15 & 34.6 \\ \hline
Ours & \underline{9.88} & 101.16 & \textbf{41.1} \\ \bottomrule
\end{tabular}%
}
\label{tab:cost}
\end{table}

\subsection{Computational Cost}
We introduce a novel tracking approach based on a decoupled architecture and compare its computational cost with existing methods (DVIS, DVIS++, and DVIS-DAQ) in \tabref{tab:cost}. All models share the same segmentation network, with difference arising only in their tracking networks. The baseline model, DVIS, has a tracker with 9.68M parameters, while DVIS++ doubles the number of channels in its attention operations, increasing its parameter count to 26.78M. Similarly, DVIS-DAQ doubles the number of transformer blocks, resulting in 18.62M parameters. In contrast, our method maintains a nearly identical parameter count to the baseline at 9.88M. 

Inference speed (measured in ms/frame) is evaluated using the R50 backbone on an RTX 2080 Ti GPU. Compared to DVIS and DVIS++, our method requires approximately 10ms and 3ms more per frame, respectively, while achieving substantial performance gains of +7.2 AP and +3.9 AP. Compared to DVIS-DAQ, our method is approximately twice as fast while obtaining +6.5 AP. These results highlight that our method achieves an effective balances performance and computational efficiency.


\section{Conclusion}
In this paper, we present \titlename~for temporally consistent video instance segmentation, a robust method designed to maintain temporal consistency in VIS. At the heart of \titlenameabb~is the \memory, which utilizes the most recent instance features to effectively discern and track objects over time. The memory is essential for correctly assigning indices to new objects and distinguishing them from existing ones.
We also propose the \matching~strategy, integrating our object memory with occupancy information to ensure precise object matching. This approach divides the object association process into two phases: continuously tracking existing objects to preserve their identities and aligning all current frame instances via an adaptive anchor query. This ensures both stable and accurate tracking.
Extensive experiments confirm that our \titlenameabb~achieves top performance across the most standardized VIS benchmarks.

\subsubsection*{Acknowledgments}
This work was supported by the Ministry of Trade, Industry and Energy (MOTIE), South Korea, through the Technology Innovation Program under Grant RS-2024-00445759, for the development of navigation technology utilizing visual information based on vision-language models for understanding dynamic environments in nonlearned spaces), the National Research Foundation of Korea (NRF) grant funded by the Korea government (MSIT) (No. RS-2023-00210908) and the Institute of Information \& Communications Technology Planning \& Evaluation(IITP) grant funded by the Korea government(MSIT) (No. RS-2025-02219277, AI Star Fellowship Support Project(DGIST)).

{
    \small
    \bibliographystyle{ieeenat_fullname}
    \bibliography{main}
}

\clearpage
\appendix

\section{Appendix}

\begin{strip}
    \centering
    \includegraphics[width=1.0\linewidth]{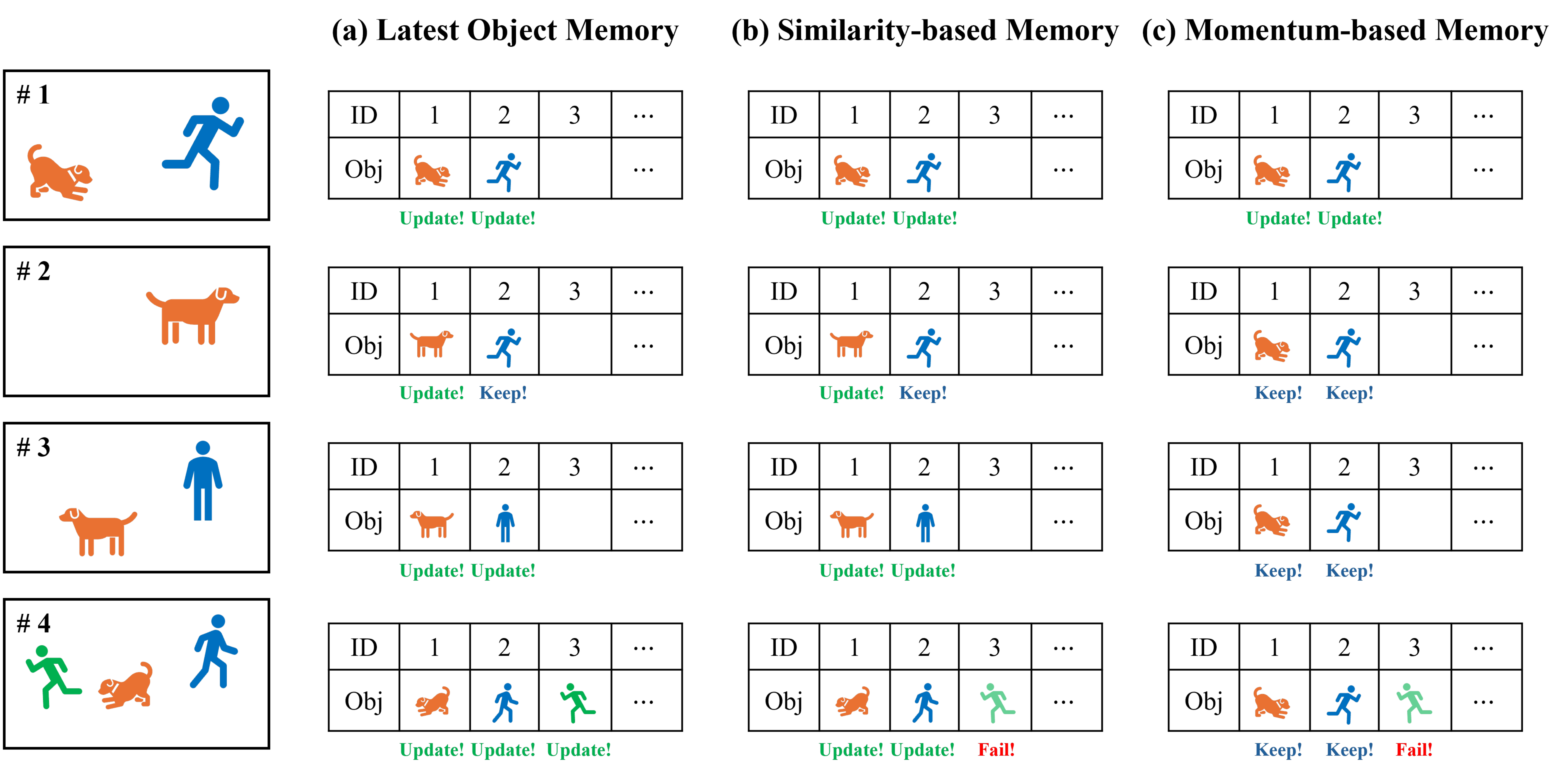}
    \includegraphics[width=1.0\linewidth]{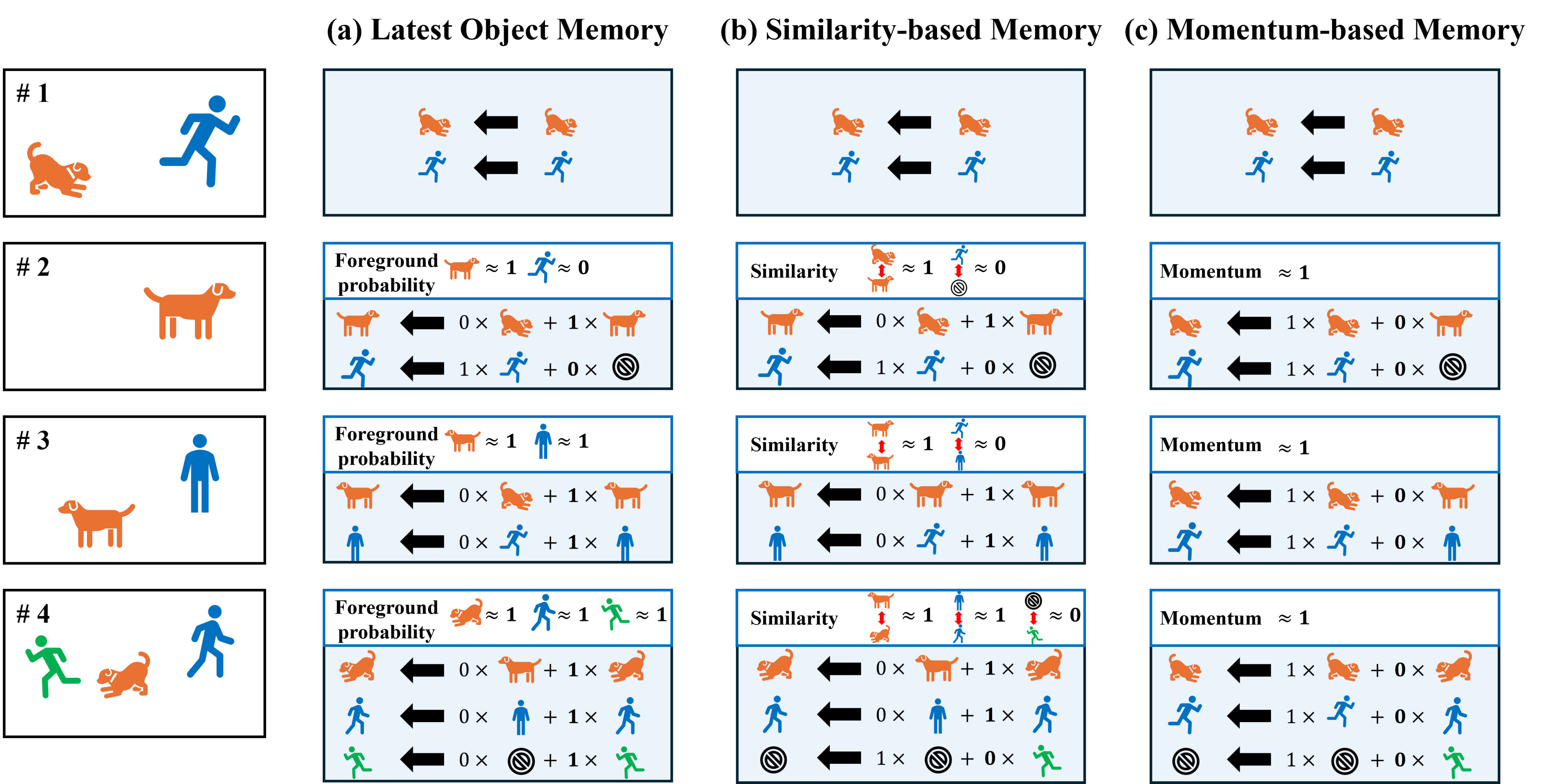}
    \captionof{figure}{Comparison of three different memory mechanisms}
    \label{fig:mem_result}
\end{strip}


\subsection{Motivation for Memory Design}
\label{apdx:motivation}
In this section, we outline the motivation behind the design of \titlenameabb, comparing three different memory systems: (1) \memory~(\memoryabb), (2) Similarity-based memory, and (3) Momentum-based memory. \figref{fig:mem_result} illustrate how each memory mechanism operates.

\textbf{Latest-state-aware object memory} uses the foreground probability of objects as a weight to update the object information at each frame. This means that the most recent and valid information is updated, allowing accurate memory updates even in scenarios where new objects appear, as seen in \figref{fig:mem_result}-(a) at frame $\# 4$. Additionally, when objects disappear, the foreground probability is low, so the previous information is largely preserved, maintaining the object information well even in situations like frame $\# 2$. This novel mechanism ensures consistent memory updates, both for newly emerging and existing objects, offering robust performance in dynamic scenarios.

\textbf{Similarity-based memory} \citep{wu2022defense, ying2023ctvis} updates the memory by assessing the similarity between the memory from time $t-1$ and the objects at time $t$. When the object information at time $t$ shows high similarity to the memory objects, it receives a significant weight during the update; however, if the similarity is low, the update is minimal. This behavior is evident in \figref{fig:mem_result}-(b), where at frame $\# 2$, the memory retains information about the person who has disappeared, while the latest information for the dog is updated. As a result, we observe a performance improvement over the baseline in Tab. \tabref{tab:ablation}-(b). However, at frame $\# 4$, when a new person appears, the memory structure struggles to update effectively due to the low similarity to the existing memory.

\textbf{Momentum-based memory} \citep{gao2023memotr} assigns a high weight (\textit{e.g.}, 99\%) to the memory information from time $t-1$ and a low weight (\textit{e.g.}, 1\%) to the objects at time $t$ during the update. As a result, the object information from the first frame is retained at a high ratio, while new objects or the latest information are hardly updated.

\subsection{Training Details}
\label{sec:a_training}
\subsubsection{Datasets.}
We evaluate the performance of our \titlenameabb~using standard benchmark datasets: YouTubeVIS datasets (2019, 2021, 2022) \citep{yang2019video} and OVIS \citep{qi2022occluded}, as detailed below.
Introduced by \citep{yang2019video} alongside the pioneering study on the Video Instance Segmentation (VIS) task, the YouTube-VIS datasets consist of high-resolution YouTube videos across 40 categories. The 2019 release includes 2,238 videos for training, 302 for validation, and 343 for testing. In its 2021 update \citep{vis2021}, the dataset was expanded to include 2,985 training videos, 421 validation videos, and 453 test videos, allowing for more extensive testing and development of VIS models. The 2022 version includes an additional 71 long videos in the validation set, while the training set remained the same as in the 2021 version.
OVIS dataset \citep{qi2022occluded} presents significant challenges with videos that often feature occlusions and long sequences that mirror complex real-world scenarios. This dataset is particularly demanding, with a greater number of objects and frames compared to YouTube-VIS, enhancing the difficulty of segmentation and tracking tasks. OVIS comprises 607 training videos, 140 validation videos, and 154 test videos, providing a robust platform for evaluating the effectiveness of VIS approaches under challenging conditions.

\subsubsection{Implementation Details.}
For our segmentation network, we employ the Mask2Former architecture \citep{cheng2022masked} equipped with three distinct backbone encoders: ResNet-50 \citep{he2016deep}, ViT-L and ViT-H \citep{dosovitskiy2021an}. 
All backbones are initialized with parameters pre-trained on COCO \citep{lin2014microsoft}.
To improve memory efficiency with the ViT-L and ViT-H, we incorporate a memory-optimized VIT-Adapter \citep{chen2022vision}, aligning with recent advancements in network efficiency \citep{zhang2023dvis++}. The segmentation network is further enhanced through pretraining with a contrastive learning approach for better object representation \citep{wu2022defense, ying2023ctvis, zhang2023dvis++, lee2024context}.
Our tracking framework integrates two networks $\mathcal{T}_E$ and $\mathcal{T}_A$, each comprising three transformer blocks and enhanced with a referring cross-attention layer \citep{DVIS} for improved accuracy.

\begin{figure*}[t!]
    \centering
    \includegraphics[width=1.0\linewidth]{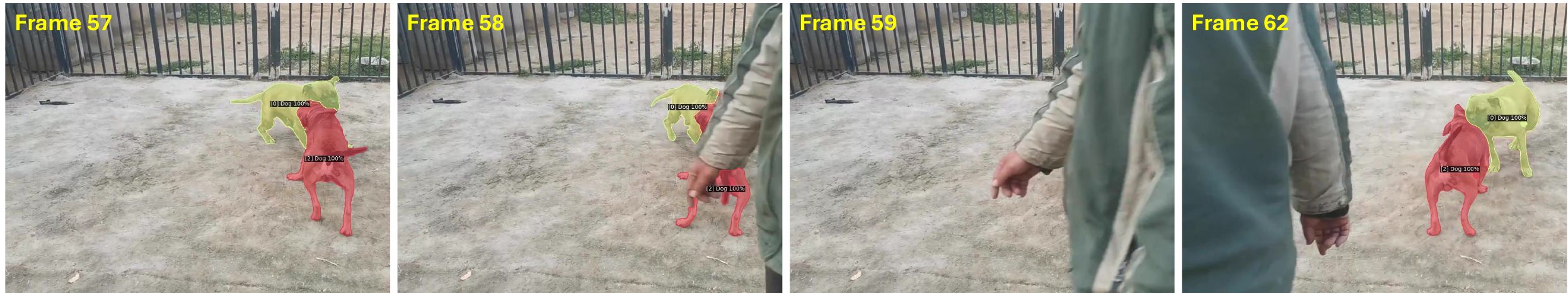}
    \caption{\textbf{Limitation.} Our method cannot handle cases where the segmentation network fails to detect objects. However, for detected objects, it demonstrates consistent tracking performance.}
    \label{fig:LOMM_limitations}
\end{figure*}

For training, our tracking networks are trained with all other parameters frozen as previous studies \citep{DVIS, li2023tcovis}. 
We employ the AdamW optimizer \citep{loshchilov2017decoupled}, initializing with a learning rate of 1e-4 and a weight decay of 5e-2.
Training is conducted over 160k iterations, with learning rate reductions scheduled at the 112k mark. We process five frames from each video in a batch of eight during training, adjusting the frame sizes to maintain a shorter side between 320 and 640 pixels, and ensuring the longer side does not exceed 768 pixels.
In all experimental settings, we incorporate COCO joint training, as utilized in prior works \citep{wu2022seqformer, heo2022vita, heo2023generalized, ying2023ctvis, DVIS}.
For inference, the shorter side of input frames is scaled to 480 pixels, maintaining uniform aspect ratios.
We adopt a temporal refiner \cite{DVIS} for our offline model.
We empirically set $\lambda_{\text{sim}}$ as 1.0.
In the online experiments using the R50 and ViT-L backbones, eight RTX2080 Ti GPUs are employed. For the offline experiments, eight RTX3090 Ti GPUs are used, while the experiments utilizing the ViT-H backbone are conducted with eight RTX A6000 GPUs.

\begin{figure*}[t!]
    \centering
    \includegraphics[width=1.0\linewidth]{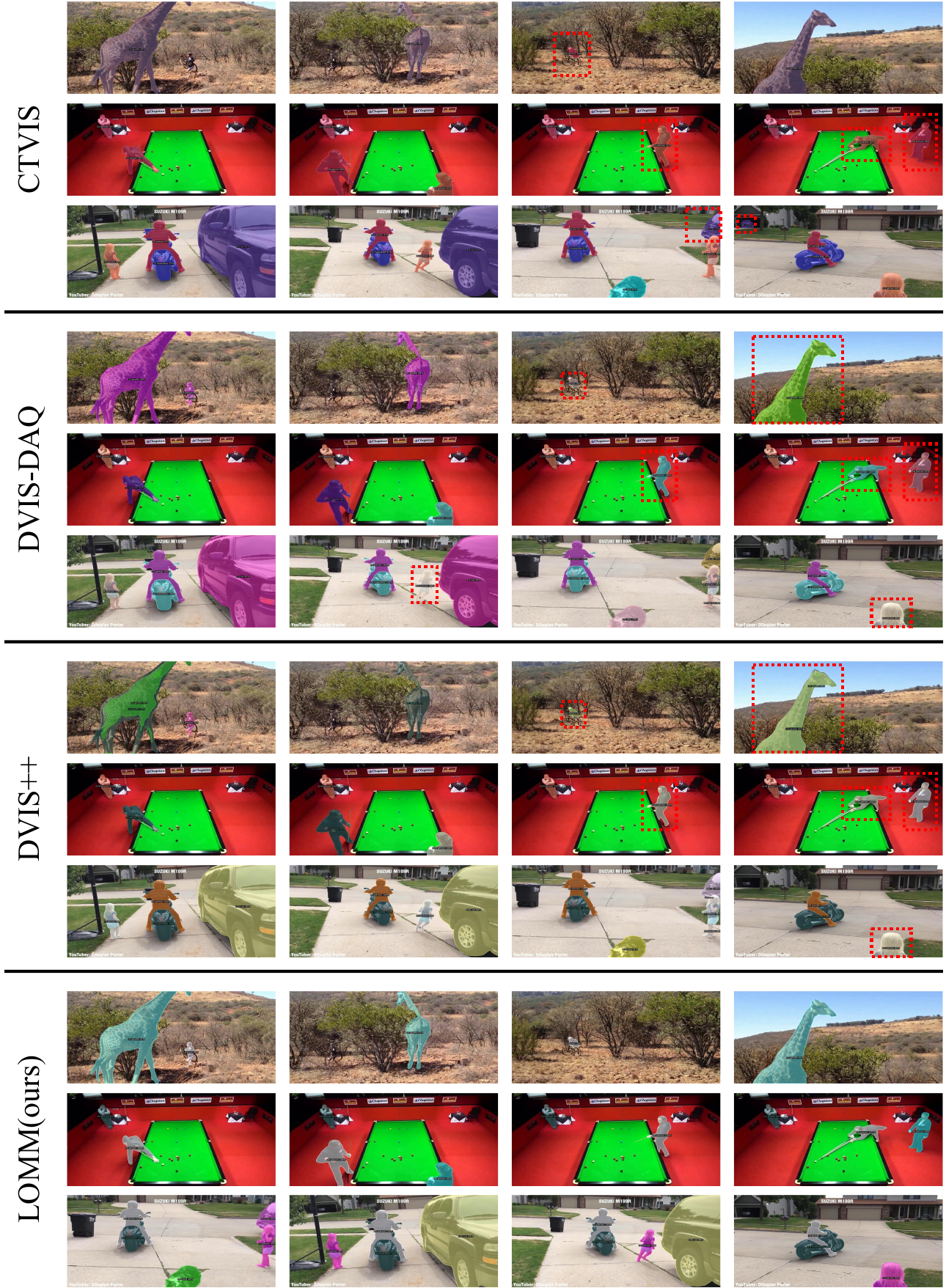}
    \caption{Qualitative comparison of \titlenameabb~with CTVIS, DVIS-DAQ, and DVIS++ on challenging scenarios in YTVIS22 dataset.}
    \label{fig:OTCVIS_comparing}
\end{figure*}

\textbf{Segmentation network.}
To achieve distinctive object representation, we employ the following contrastive loss for pretraining the segmentation network $\mathcal{S}$:
\begin{equation}
\begin{aligned}
\mathcal{L}_{\text {embed }} & =-\log \frac{\exp \left(\mathbf{v} \cdot \mathbf{k}^{+}\right)}{\exp \left(\mathbf{v} \cdot \mathbf{k}^{+}\right)+\sum_{\mathbf{k}^{-}} \exp \left(\mathbf{v} \cdot \mathbf{k}^{-}\right)} \\
&=\log \left[1+\sum_{\mathbf{k}^{-}} \exp \left(\mathbf{v} \cdot \mathbf{k}^{-}-\mathbf{v} \cdot \mathbf{k}^{+}\right)\right],
\end{aligned}
\end{equation}
where $\mathbf{k^+}$, and $\mathbf{k^-}$ denote positive embedding and negative embedding from anchor embedding $\mathbf{v}$.
This contrastive loss is widely applied in the VIS field \citep{wu2022defense, li2023mdqe, ying2023ctvis, zhang2023dvis++, lee2024context}, learning frame-to-frame associations to create better object representations.

\textbf{Early training.}
The initial outputs from the tracking networks $\mathcal{T}_{E}$ and $\mathcal{T}_{A}$ are also typically noisy. 
To address this, we utilize the predictions $\hat{y}$ from $\tilde{Q}^*_{t}$ for ground truth assignment, formulated as:
\begin{equation}
\label{eq:matching}
    \begin{gathered}
        \acute{\sigma} = \underset{\sigma\in \mathfrak{S}_{N}}{\arg \min } \sum_{n=1}^{N_{GT}} \mathcal{L}_{\text {Match}}\left(y_{f(n)}^n, \hat{y}_{f(n)}^{\sigma(n)}\right).
    \end{gathered}
\end{equation}
The prediction $\hat{y}$ provides guidance for rapid convergence in the same format as the tracked output of MinVIS \citep{huang2022minvis}.

\subsubsection{Additional Qualitative Results}
\label{sec:a_qual}
We provide additional comparisons with state-of-the-art models, as shown in \figref{fig:OTCVIS_comparing}, to highlight the robustness of our model in challenging scenarios where objects frequently appear and disappear. Existing methods, including CTVIS \cite{ying2023ctvis}, DVIS-DAQ \cite{zhou2024dvis}, and DVIS++ \cite{zhang2023dvis++}, often fail to track accurately by either misidentifying reappearing objects as new or confusing newly appeared objects with existing ones. By utilizing a robust memory mechanism and an effective object association strategy, our model maintains consistently discriminative embeddings, significantly enhancing both segmentation and tracking performance.

\subsubsection{Limitation}

Our method adopts a decoupled framework, where the segmentation network is frozen while training the tracking network. As a result, it cannot handle objects that the segmentation network fails to detect. Nevertheless, our approach achieves significant improvements in long-term consistent tracking, a fundamental challenge in VIS. As shown in \figref{fig:LOMM_limitations}, even when the segmentation network misses certain objects, our method maintains robust tracking.

\end{document}